# Combined Descriptors in Spatial Pyramid Domain for Image Classification

Junlin Hu and Ping Guo

*Image Processing and Pattern Recognition Laboratory*
*Beijing Normal University, Beijing 100875, China*
`hujunlin@msn.com,pguo@ieee.org`

**Abstract.** Recently spatial pyramid matching (SPM) with scale invariant feature transform (SIFT) descriptor has been successfully used in image classification. Unfortunately, the codebook generation and feature quantization procedures using SIFT feature have the high complexity both in time and space. To address this problem, in this paper, we propose an approach which combines local binary patterns (LBP) and three-patch local binary patterns (TPLBP) in spatial pyramid domain. The proposed method does not need to learn the codebook and feature quantization processing, hence it becomes very efficient. Experiments on two popular benchmark datasets demonstrate that the proposed method always significantly outperforms the very popular SPM based SIFT descriptor method both in time and classification accuracy.

## 1 Introduction

Image classification is a basic problem in computer vision for decades and has attracted many researchers' attention these years. Many image representation models have been proposed for this problem, such as probabilistic latent semantic analysis (pLSA) [1], part-based model [2], bag of visual words (BoW) model [3], etc. From these existing methods, BoW model has shown excellent performance and been widely used in many real applications due to its robustness to scale, translation and rotation variance, including image classification [4], image annotation [5], image retrieval [6] and video event detection [7]. The BoW model treats an image as a collection of unordered appearance descriptors extracted from local patches, quantizes them into discrete visual words, and then computes a compact histogram representation for image classification.

However, the BoW approach discards the spatial information of local descriptors, the descriptors representation power is severely limited. To include information about the spatial layout of the features, an extension of BoW model named as spatial pyramid matching (SPM) [8] was proposed, and achieved excellent results on the several challenging image classification tasks. For each image, SPM estimates the overall perceptual similarity between images, which can be used as the kernel function in support vector machines (SVM). Recently, many extension of SPM have also been developed by sparse coding [9] and multiple pooling strategy [10].

Though SIFT descriptors with SPM has achieved good performance on image classification, codebook generation and feature quantization are the most important and these procedures have a high complexity both in time and space. To address this problem, multi-level kernel machine [11] and spatial local binary patterns (SLBP) [12] were introduced recently. Unlike these approaches, in this paper, we adopt local binary patterns descriptor with SPM, and three-patch LBP descriptor with SPM for image classification, respectively. We also propose to combine LBP and TPLBP in spatial domain for further improving the discriminative power of descriptors. Because our proposed methods does not need to learn the codebook and feature quantization procedures, hence it becomes very efficient. Experiments on 15 class scene category dataset and Caltech 101 dataset show that our method outperforms the very popular SPM based SIFT method both in time and classification accuracy.

The rest of the paper is organized as follows. Section 2 briefly introduces LBP, TPLBP and SPM, respectively. Section 3 gives our proposed combined approach in spatial pyramid domain. Section 4 presents experiment setup and results on two public datasets. Finally, we conclude our paper in Section 5.

## 2   Related Work

### 2.1   LBP and TPLBP

Local binary patterns (LBP) introduced by Ojala *et al.* [13], is invariance against monotonic gray-scale changes, low computational complexity and convenient multi-scale extension. There are many extensions to the original LBP descriptor, including uniform patterns [14], Center-Symmetric LBP (CSLBP) [15], Three-Patch LBP (TPLBP) [16], *etc.* In our work, we only extract LBP and TPLBP descriptors. The TPLBP is briefly described as following.

The TPLBP code is produced by comparing the values of three patches to produce a single bit value in the code assigned to each pixel. For each pixel in the image, we consider a windows of $w \times w$ patch centered on the pixel, and $S$ is the total number of windows additional patches distributed uniformly in a ring of radius $r$ around it (seen in Fig. 1). For a parameter $\alpha$, we take pairs of patches, $\alpha$-patches apart along the circle, and compare their values with the central patch. The value of a single bit is set according to which of the two patches is more similar to the central patch. The resulting code has $S$ bits per pixel. The TPLBP code to each pixel is presented by the following formula:

$$\text{TPLBP}_{r,S,w,\alpha}(p) = \sum_{i=1}^{S-1} f(d(C_i, C_p) - d(C_{i+\alpha \bmod S}, C_p))2^i \qquad (1)$$

Where $C_i$ and $C_{i+\alpha \bmod S}$ are two patches along the ring and $C_p$ is the central patch. The function $d(\cdot, \cdot)$ is any distance function between two patches and $f$

is defined as:

$$f(x) = \begin{cases} 1, & \text{if } x \geq \tau \\ 0, & \text{if } x < \tau \end{cases} \quad (2)$$

Here, $\tau$ (*e.g.*, $\tau = 0.01$) is slightly greater than zero to provide some stability in uniform regions.

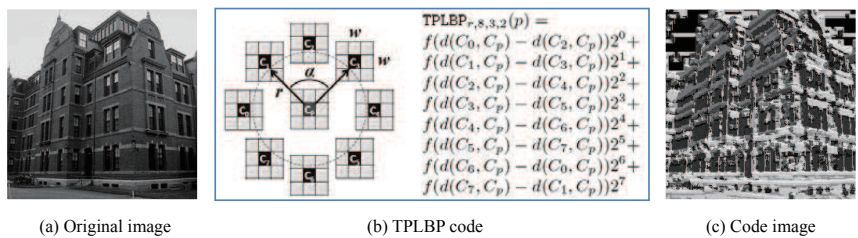

(a) Original image    (b) TPLBP code    (c) Code image

**Fig. 1.** TPLBP descriptors. (a) Original image. (b) The TPLBP code computed with parameters $S = 8$, $w = 3$, and $\alpha = 2$. (c) Code image produced from original image.

### 2.2 Spatial Pyramid Matching

The pyramid matching framework was proposed by Lazebnik *et al.* [8] to find an approximate correspondence between two sets of vectors in a feature space. Generally, pyramid matching works by placing a sequence of increasingly coarser grids over the image and taking a weighted sum of the number of matches that occur at each scale. Feature matches from finer scales are given more weight. In that, spatial pyramid matching works in $L$ levels of image resolutions, and in level $l$, the image is partitioned to $2^l \times 2^l$ grids of the same size.

For two images $I_1$ and $I_2$, spatial pyramid matching kernel $K$ is defined as:

$$K(I_1, I_2) = \sum_{l=0}^{L} \sum_{j=1}^{J_l} w_{l,j} K_{l,j}(I_1, I_2) \quad (3)$$

Here $L$ is the total number of levels and $J_l$ is the total number of grids in level $l$, $w_{l,j}$ is the weight for the $j^{th}$ grid in the level $l$. In [8], it is chosen as:

$$w_{l,j} = \begin{cases} \frac{1}{2^L}, & \text{if } l = 0 \\ \frac{1}{2^{L-l+1}}, & \text{if } l > 0 \end{cases} \quad (4)$$

$K_{l,j}(I_1, I_2)$ is the number of matches at level $l$, which is given by the histogram intersection function:

$$K_{l,j}(I_1, I_2) = \sum_{m=1}^{M} \min(H_{l,j}(I_1), H_{l,j}(I_2)) \quad (5)$$

$H_{l,j}(I_1)$ is the histogram where $m$ is gray levels appearing in $j^{th}$ grid of $l^{th}$ level in image $I_1$. In practice, it is reported that $L = 2$ or $L = 3$ is enough [17].

## 3    Proposed Approach

The LBP and TPLBP descriptors have $M$ discrete channels. Then, spatial pyramid representation partitions an image into segments in different scales, and computes the histogram of each segment, and finally concatenates all the histograms to form a vector representation of the image (shown in Fig. 2). Having spatial pyramid representation for each image, SPM kernel is used to estimate the similarity between images. Lastly, support vector machines (SVM) [18] with spatial pyramid matching kernel $K$ is used to perform image classification tasks.

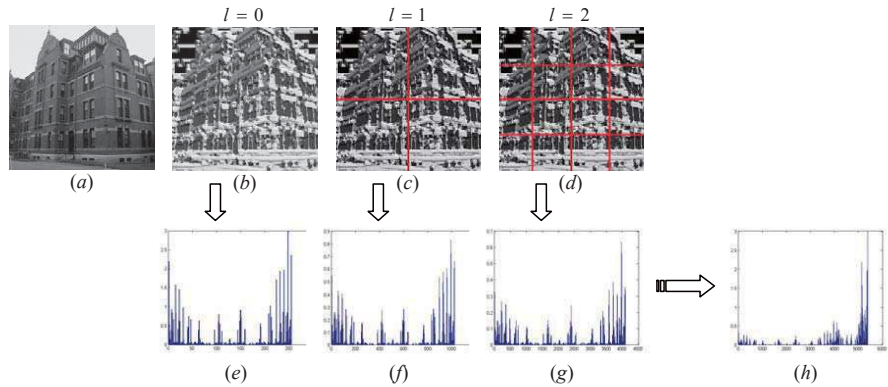

**Fig. 2.** TPLBP (or LBP) descriptor extraction in spatial pyramid domain. (a) Original image. (b) (c) (d) The grids on the TPLBP code image for each level $l = 0, 1, 2$, respectively. (e) (f) (g) Histogram representations corresponding to (b) (c) (d), respectively. (h) The final TPLBP vector (TPLBPSPM) is a weighted concatenation of histograms for all levels.

What's more, we also propose to combine LBP and TPLBP in spatial pyramid domain for further improving the discriminative power of descriptors. For simplicity, LBPSPM denotes LBP in SPM domain, similar to TPLBPSPM, the combined approach is abbreviated as ComSPM, which can be written as:

$$\text{ComSPM} = \lambda \text{ LBPSPM} + (1 - \lambda) \text{ TPLBPSPM} \quad (6)$$

Here, $\lambda$ ($0 \leq \lambda \leq 1$) is a weighted parameter. By tuning the parameter $\lambda$, we can investigate the relationship between LBPSPM and TPLBPSPM, and discover how $\lambda$ effects our ComSPM method and where our approach can achieve the best performance. The flowchart of our ComSPM method is shown in Fig. 3.

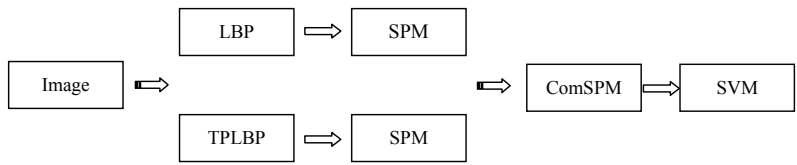

**Fig. 3.** The flow chart of the proposed ComSPM approach.

## 4 Experiments

In this section, we evaluate our method on two widely used datasets: the 15 class scene category [8] and Caltech 101 dataset [19]. Following the same experiment procedure of SPM based SIFT descriptors [8], the SIFT descriptors are densely sampled from each image on the $16 \times 16$ pixels patch located every 8 pixels. To fairly compare with others, we fixed the spatial pyramid levels $L = 2$ and channels $M = 256$ for all the tests. Experiments are repeated ten times with different randomly selected training and test images to obtain statical results. The finally result are shown as the mean and standard deviation of the recognition rates. Noted that we empirically fixed $\lambda = 0.3$ for all the tests.

### 4.1 The 15 class scene category dataset

The 15 class scene category dataset contains totally 4485 images with 15 categories (including coast, forest, kitchen, office, *etc.*). Images are about $300 \times 250$ in average size, with 210 to 410 images in each category. Fig. 4 gives fifteen representative example images from this dataset.

Following the same experiment procedure of the SIFT descriptors with SPM [8], we took 100 images per class for training and used the left for testing. The comparison results are shown in Table 1. Moreover, Fig. 5 shows a confusion table between the fifteen scene categories using our ComSPM approach.

Table 1 shows that our LBPSPM and TPLBPSPM is slightly lower than SPM based SIFT about 3% and 2.6% in classification rate, respectively. However, the average processing time for our method in generating the final representation from a raw image input is only 0.11 second and 0.45 second separately. Moreover, we notice that our ComSPM method outperforms SPM based SIFT descriptor by nearly 1.5% in recognition rate and has a significant advantage of saving time (reduced by nearly 2.29 times).

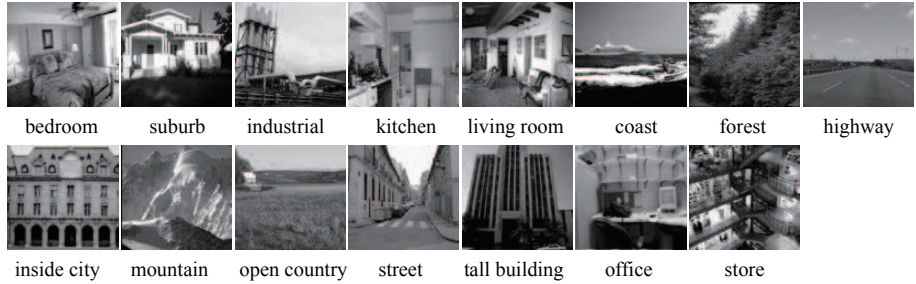

bedroom   suburb   industrial   kitchen   living room   coast   forest   highway

inside city   mountain   open country   street   tall building   office   store

**Fig. 4.** Example images from the 15 class scene category dataset. One per category.

**Table 1.** Time (seconds per image) and Classification rate (%) comparison on the 15 class scene category dataset.

| Method | $M$ | Time ($seconds$) | Classification Rate (%) |
|---|---|---|---|
| SIFTSPM [8] | 256 | 1.28 | $81.38 \pm 0.24$ |
| LBPSPM | 256 | **0.11** | $78.34 \pm 0.31$ |
| TPLBPSPM | 256 | **0.45** | $78.70 \pm 0.32$ |
| ComSPM | 256 | **0.56** | $\mathbf{82.68 \pm 0.25}$ |

|  | suburb | coast | forest | highway | inside city | mountain | open country | street | tall building | office | bedroom | industrial | kitchen | living room | store |
|---|---|---|---|---|---|---|---|---|---|---|---|---|---|---|---|
| suburb | 98.58 | 0 | 0 | 0 | 0 | 0 | 0 | 0 | 0 | 0 | 0.71 | 0 | 0 | 0.71 | 0 |
| coast | 0 | 91.54 | 0.77 | 1.54 | 0 | 0.77 | 5.38 | 0 | 0 | 0 | 0 | 0 | 0 | 0 | 0 |
| forest | 0 | 0 | 90.79 | 0 | 0 | 7.02 | 1.32 | 0 | 0.44 | 0 | 0 | 0.44 | 0 | 0 | 0 |
| highway | 0 | 6.25 | 0 | 85.62 | 1.25 | 1.87 | 3.75 | 0.63 | 0 | 0 | 0 | 0 | 0 | 0 | 0.63 |
| inside city | 0 | 0.48 | 0 | 0.96 | 87.02 | 0 | 0 | 3.85 | 7.69 | 0 | 0 | 0 | 0 | 0 | 0 |
| mountain | 0 | 3.28 | 4.38 | 0.73 | 0 | 84.67 | 5.84 | 0.36 | 0.73 | 0 | 0 | 0 | 0 | 0 | 0 |
| open country | 0 | 13.87 | 4.84 | 2.58 | 0.32 | 8.39 | 69.68 | 0.32 | 0 | 0 | 0 | 0 | 0 | 0 | 0 |
| street | 0 | 0 | 0 | 2.08 | 5.21 | 1.04 | 0.52 | 86.98 | 4.17 | 0 | 0 | 0 | 0 | 0 | 0 |
| tall building | 0 | 0 | 0.39 | 1.17 | 4.30 | 0.78 | 0.39 | 0.78 | 92.19 | 0 | 0 | 0 | 0 | 0 | 0 |
| office | 0 | 0 | 0 | 0 | 0 | 0 | 0 | 0 | 0 | 96.52 | 0.87 | 0 | 1.74 | 0.87 | 0 |
| bedroom | 0 | 0 | 0 | 0 | 0 | 0 | 0 | 0 | 0 | 6.90 | 55.17 | 6.03 | 5.17 | 24.14 | 2.59 |
| industrial | 1.90 | 0 | 0 | 0 | 0 | 0 | 0 | 0 | 0 | 0 | 3.79 | 80.57 | 1.90 | 0.47 | 11.37 |
| kitchen | 0 | 0 | 0 | 0 | 0 | 0 | 0 | 0 | 0 | 3.64 | 4.55 | 6.36 | 75.45 | 5.45 | 4.55 |
| living room | 0.53 | 0 | 0 | 0 | 0 | 0 | 0 | 0 | 0 | 7.94 | 21.69 | 0.53 | 4.76 | 60.32 | 4.23 |
| store | 1.40 | 0 | 0 | 0 | 0 | 0 | 0 | 0 | 0 | 0.47 | 0.93 | 11.63 | 5.58 | 0.93 | 79.07 |

**Fig. 5.** The Confusion Matrix on the 15 class scene category dataset (%). Average classification rates for individual classes are listed along the diagonal. The entry in the $i^{th}$ row and $j^{th}$ column is the percentage of images from class $i$ that are misidentified as class $j$. Here, average classification rate is 82.58%.

From the confusion matrix in Fig. 5, it can be observed that, our method performs better for category suburb, coast, forest, tall building and office (more than 90%). Furthermore, we notice that the category bedroom, living room, kitchen and open country have a high percentage being classified wrongly. Not surprisingly, this may result from that they are visually similar to each other.

### 4.2  Caltech 101 dataset

Our second dataset is the Caltech 101 database [19]. This dataset collects 101 classes (including animals, vehicles, flowers, etc.) with high shape variability. The number of images per category varies from 31 to 800. Most images are medium resolution (about $300 \times 250$ pixels).

**Table 2.** Classification rate (%) comparison on Caltech 101 dataset.

| Method | M | Classification Rate (%) |
|---|---|---|
| SIFTSPM [8] | 256 | $64.06 \pm 0.50$ |
| LBPSPM | 256 | $58.57 \pm 0.41$ |
| TPLBPSPM | 256 | $63.60 \pm 0.38$ |
| ComSPM | 256 | **$65.50 \pm 0.49$** |

We followe the common setup during experiment, that is, training on 30 images per category and testing on the rest. The detailed comparison results on this dataset are listed in Table 2. From this table we can see that our TPLBPSPM performs as well as the SIFT descriptors with SPM, moreover, our ComSPM method outperforms the SIFT with SPM by about 1.5 percent. Besides, for our methods, the average time in processing each image is much less than the SPM based SIFT descriptors [8] (For detailed time comparison results, we can refer to the average time per image exhibited in Table 1.).

## 5  Conclusion

In this paper, we proposed a promising combined approach by combining LBP and TPLBP in spatial pyramid domain for image classification. Unlike the most popular methods in BoW model, our method does not need to the codebook generation and feature quantization procedures, hence it becomes very efficient. Experimental results on two public datasets show that the proposed methods achieve excellent performance both in time and classification accuracy. All these sufficiently demonstrate the effectiveness of our method. In the future, we plan to advance the study by including more data and other datasets. Moreover, we may adopt sparse coding scheme in our method to further improve classification accuracy.

8       J. Hu & P. Guo

**Acknowledgement.** The research work described in this paper was fully supported by the grants from the National Natural Science Foundation of China (Project No.90820010). Prof. Ping Guo is the author to whom all the correspondence should be addressed, his e-mail address is pguo@ieee.org.